\documentclass[times, review, 10pt]{elsarticle}

\usepackage{amsmath,amssymb,amsfonts,color}
\usepackage{graphicx}
\usepackage{textcomp}
\usepackage{xcolor}
\usepackage[dvipsnames]{xcolor}
\usepackage{enumerate}
\usepackage{setspace}
\usepackage{bm}
\usepackage{comment}
\usepackage{caption}
\usepackage{tipa}
\usepackage{footnote}
\usepackage{subfigure}
\usepackage{algorithm}
\usepackage{algorithmic}
\usepackage{marvosym}
\usepackage{lineno}
\usepackage{rotating}

\usepackage{multirow}
\usepackage{multicol}
\usepackage{subfigure}
\usepackage{multirow}
\usepackage{url}
\usepackage{booktabs}
\usepackage{tabularx}
\usepackage{adjustbox}
\usepackage{arydshln}

\def\x{{\mathbf x}}

\usepackage{url}
\usepackage{hyperref}
\newcolumntype{C}[1]{>{\centering\arraybackslash\vspace{0pt}}m{#1}}
\newcolumntype{Y}{>{\centering\arraybackslash}X}
\newcolumntype{P}{>{\centering\arraybackslash}p}

\begin{document}

\begin{frontmatter}

\title{Difficulty-guided Sampling: Bridging the Target Gap between Dataset Distillation and Downstream Tasks}
\author{Mingzhuo Li${}^\text{a}$}
\ead{mingzhuo@lmd.ist.hokudai.ac.jp}
\author{Guang Li${}^\text{a}$}
\ead{guang@lmd.ist.hokudai.ac.jp}
\author{Linfeng Ye${}^\text{b}$}
\ead{linfeng.ye@mail.utoronto.ca}
\author{Jiafeng Mao${}^\text{c}$}
\ead{mao@hal.t.u-tokyo.ac.jp}
\author{Takahiro Ogawa${}^\text{a}$}
\ead{ogawa@lmd.ist.hokudai.ac.jp}
\author{Konstantinos N. Plataniotis${}^\text{b}$}
\ead{kostas@ece.utoronto.ca}
\author{Miki Haseyama${}^\text{a}$}
\ead{mhaseyama@lmd.ist.hokudai.ac.jp}
\address{${}^\text{a}$Hokkaido University, N-14, W-9, Kita-Ku, Sapporo, 060-0814, Japan}
\address{${}^\text{b}$University of Toronto, 27 King's College Circle, Toronto, Ontario M5S 1A1, Canada}
\address{${}^\text{c}$The University of Tokyo, 7-3-1 Hongo, Bunkyo-ku, Tokyo, 113-8654, Japan}

\doublespacing

\begin{abstract}
In this paper, we propose difficulty-guided sampling (DGS) to bridge the target gap between the distillation objective and the downstream task, therefore improving the performance of dataset distillation. Deep neural networks achieve remarkable performance but have time and storage-consuming training processes. Dataset distillation is proposed to generate compact, high-quality distilled datasets, enabling effective model training while maintaining downstream performance. Existing approaches typically focus on features extracted from the original dataset, overlooking task-specific information, which leads to a target gap between the distillation objective and the downstream task. We propose leveraging characteristics that benefit the downstream training into data distillation to bridge this gap.
Focusing on the downstream task of image classification, we introduce the concept of difficulty and propose DGS as a plug-in post-stage sampling module. Following the specific target difficulty distribution, the final distilled dataset is sampled from image pools generated by existing methods. We also propose difficulty-aware guidance (DAG) to explore the effect of difficulty in the generation process. Extensive experiments across multiple settings demonstrate the effectiveness of the proposed methods. It also highlights the broader potential of difficulty for diverse downstream tasks.
\end{abstract}


\end{frontmatter}

\doublespacing

\section{Introduction}

\label{sec:intro}
Deep neural networks have gained widespread recognition with the rapid advancement of deep learning and the dramatic growth of information produced in daily life \cite{talaei2023review}. They are capable of automatically learning representative features, achieving remarkable performance with strong generalization ability, and have been successfully applied across various fields. However, the performance of deep neural networks heavily depends on network complexity, which in turn requires large amounts of data for effective training \cite{kaplan2020cost}. This dependence results in substantial demands for computational and storage resources, as well as prolonged training times. For instance, the ImageNet-1K dataset \cite{deng2015imageNet1k}, commonly used for training classification models such as ResNet \cite{he2016resNetAP} or ConvNet \cite{gidaris2018convNet}, exceeds 130 GB in size, requiring training times of tens of hours or even many days. 
\par
\begin{figure}[t]
    \centering
    \includegraphics[width=0.8\linewidth]{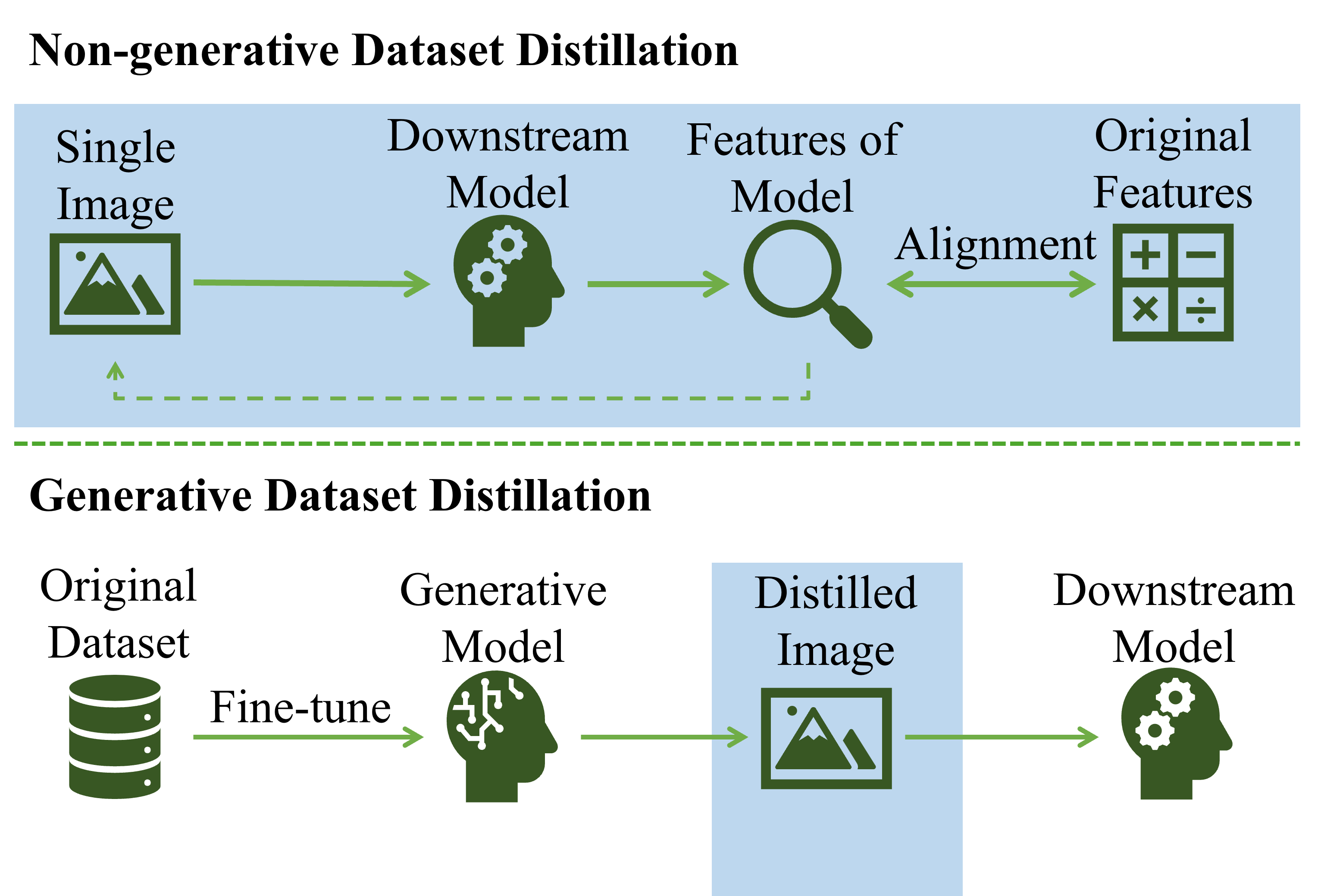}
    \caption{Illustration of the workflow of non-generative and generative dataset distillation methods, summarized from current methods. The blue areas show the part that is repeated when generating multiple images.}
    \label{intro}
\end{figure}

During the development and practical deployment of deep neural networks, several common scenarios may become annoying due to the enormous size of the datasets involved. First, the dataset may be too large for one device to process in a single run, resulting in prolonged device occupation, which is particularly problematic in constrained computing environments. To mitigate this, researchers typically adopt two strategies: dividing the dataset into chunks and performing checkpoint-based training to split the process over time, or utilizing multiple devices to accelerate training \cite{sergeev2018distribute}. However, both approaches require additional effort to ensure accurate implementation. Second, the sheer cost of one single training also hinders research that requires repeatedly training models from scratch. For example, to identify an appropriate network architecture, tune hyperparameters, or perform ablation studies \cite{poyser2024PRArchiReview}. Finally, in certain privacy-sensitive domains, additional measures such as differential privacy or federated training have to be considered to prevent data leakage while preserving model performance \cite{liu2021privacyReview}.

\par

To solve these problems, dataset distillation \cite{wang2018datasetdistillation, li2022awesome} has been proposed to simplify the training process by distilling a large original dataset into a much smaller synthetic one. The goal is for models trained on the distilled dataset to achieve comparable performance to those trained on the original dataset, thereby enabling efficient training within a significantly reduced time. Meanwhile, compressing the original dataset also helps alleviate challenges related to data storage and transfer. Moreover, since the original dataset is no longer directly accessed in the downstream task, privacy problems are also mitigated \cite{li2020soft, li2022compressed}. Since its introduction, dataset distillation has attracted considerable attention, with an increasing number of studies contributing to its rapid development \cite{liu2025DDreview}. As shown in Fig.~\ref{intro}, existing approaches can be broadly categorized into non-generative and generative methods. Non-generative methods optimize a single image by aligning specific features originated from the original training process, and repeat the optimization process to obtain the distilled dataset. In contrast, generative dataset distillation methods leverage generative models to obtain distilled datasets, significantly reducing the generation time. The distillation targets are achieved by fine-tuning the models or conducting post-stage processing. Diffusion models, as a representative type of generative models, have demonstrated high fidelity and remarkable robustness \cite{chang2026PRdiffReview}, further accelerating progress in generative dataset distillation.

\par

Recent studies on generative dataset distillation have substantially improved the performance of distilled datasets. However, existing methods primarily concentrate on preserving the features of the original dataset \cite{geng2023DDreview}, making the optimization objective to produce a high-quality representation of the original dataset. This consideration does not include task-specific information and is expected to work across various downstream tasks. However, a specific model is generally designed to tackle a particular task in practice. Moreover, different downstream tasks may prefer different types of information, requiring tailored treatment. For example, classification tasks emphasize discriminative features that effectively separate classes \cite{Singh2020classiReview}, while object detection tasks focus more on spatial and structural information that supports accurate localization and bounding box prediction \cite{tang2021PRobjDetReview}. Such differences lead to a target gap between the distillation objective and the downstream task, limiting the optimal performance of distilled datasets. To eliminate this limitation, we propose introducing task-specific information into the distillation process to bridge the target gap, obtaining distilled datasets that better fulfill the needs of the target downstream task.

\par
\begin{figure}[t]
    \centering
    \includegraphics[width=\linewidth]{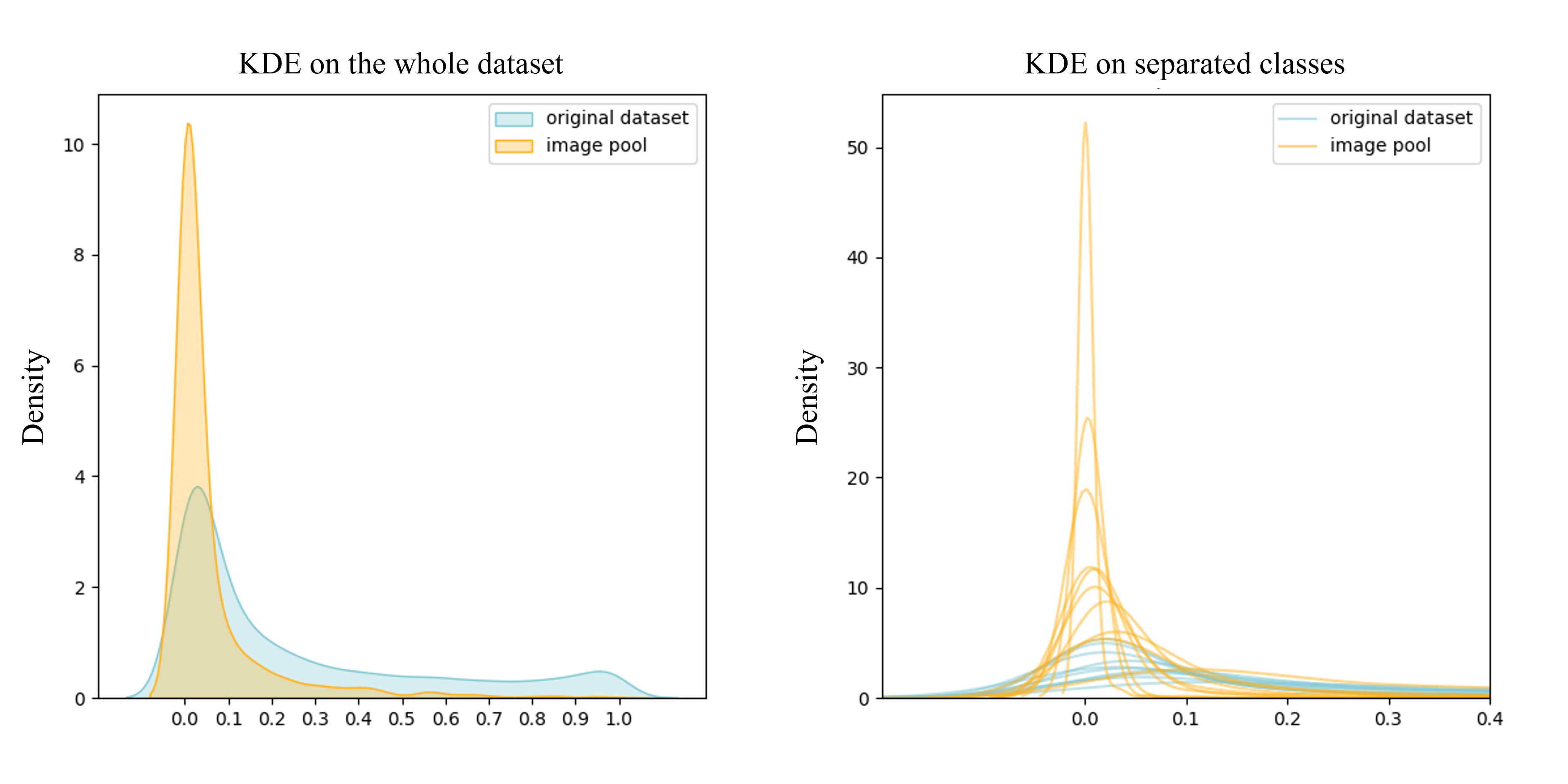}
    \caption{Kernel density estimates of the difficulty distributions for the original dataset and the image pool generated by Minimax \cite{gu2024minimax}. The left figure shows the results across the entire dataset. The second figure shows class-wise density estimates, and the horizontal axis range is adjusted to around $0 \sim 0.4$ for better readability.}
    \label{fig_kde}
\end{figure}

In this paper, focusing on the specific downstream task of image classification, we propose Difficulty-guided Sampling (DGS) to bridge the target gap, which incorporates difficulty by means of post-stage sampling. Specifically, we begin by constructing an image pool consisting of images generated by an existing generative dataset distillation method. The sampling distribution is scaled from the difficulty distribution of the original dataset. Based on the observation on distributional bias between the image pool and the original dataset shown in Fig.~\ref{fig_kde}, Distribution smoothing is introduced to align the two distributions more closely for better sampling. With these efforts, distilled datasets produced by DGS follow the original difficulty distribution, benefiting the training of classification models. We also explore an alternative solution involving difficulty-aware guidance (DAG) to directly generate datasets that inherently follow the desired distribution. Extensive experiments across multiple datasets and downstream models demonstrate that DGS and DAG consistently improve classification performance, validating the effectiveness of the proposed methods.

\par

The contributions of this paper are summarized as follows:
\begin{itemize}
    \item We propose DGS, which incorporates the task-specific information of difficulty to bridge the target gap between dataset distillation and the downstream task, improving the performance of distilled datasets on the specific task of image classification.
    
    \item We apply DGS as a plug-in post-stage sampling module on image pools obtained by existing generative dataset distillation methods. The sampling follows the difficulty distribution of the original dataset with distribution smoothing for better coverage. 
    
    \item We further introduce DAG to validate difficulty-based dataset distillation from another perspective by generating datasets with the desired difficulty distribution.
    
    \item We conduct extensive experiments across multiple datasets, downstream model architectures, and baseline methods. The results demonstrate that the proposed method consistently improves the performance, validating the effectiveness of task-specific information in dataset distillation.
\end{itemize}

\par

The paper is organized as follows: Section 2 reviews related works. Section 3 introduces the detailed implementation of GDS and further discusses the DAG. Section 4 presents the experimental results and the corresponding analysis. Section 5 summarizes key findings and insights gained during the development and evaluation of the proposed methods. Finally, Section 6 concludes the paper.

\section{Related Work}
\subsection{Generative Models}
Generative models aim to learn the underlying data distribution and synthesize new samples that resemble the original data. Early frameworks such as Variational Autoencoders (VAE) \cite{kingma2014VAE} and Generative Adversarial Networks (GAN) \cite{goodfellow2014gan} provide two representative paradigms: latent-variable modeling and adversarial learning. VAEs formulate generation through probabilistic inference by optimizing the evidence lower bound, enabling learning of structured latent representations. While GANs introduce a min–max adversarial objective that significantly improves sample realism. The strong ability of distribution modeling and high quality of generated samples contribute to the widespread use of generative models, especially in the field of computer vision.

\par

Recently, diffusion models \cite{chen2024diffReview} have received great attention for state-of-the-art (SOTA) synthesis quality and stability. They construct a forward diffusion process that gradually corrupts data with noise, and learn a reverse denoising process to recover samples, grounded in variational inference and score-matching theory. Later advances, such as DDIM \cite{song2021ddim} and latent diffusion models \cite{Rombach2022latentDiff}, greatly reduce sampling steps and extend diffusion models to large-scale and multi-modal scenarios. With high fidelity and remarkable robustness, diffusion models have become the dominant framework in modern generative models.


\subsection{Dataset Distillation}
Dataset distillation \cite{wang2018datasetdistillation} aims to construct a smaller dataset that can retain the training ability of the original dataset, allowing models trained on the distilled dataset to achieve comparable performance to those trained on the original dataset, while reducing training time and computation costs. Moreover, it also alleviates storage and transmission overhead as well as privacy benefits of avoiding direct exposure of the original data during the training process. Since its inception, dataset distillation has gained increasing attention and has been applied across various scenarios, including continual learning \cite{gu2024DDcontinue} and privacy-preserving learning \cite{zheng2025DDprivacy, li2023sharing}. Existing approaches can be broadly categorized into non-generative and generative methods.

\par

Traditional non-generative methods generally repeat the optimization on a single synthetic image to produce the distilled dataset. The repeat time equals the dataset size, usually measured by the image-per-class (IPC) \cite{li2024iadd}. The optimization is typically performed by aligning a specific feature derived from the distilled data with that from the original data. Different choices of the target feature lead to various methods. For example, DC \cite{wang2025gradMatch0} and EDF \cite{kim2022IDC} directly align the gradient dynamics, while MTT \cite{cazenavette2022MTT} and MCT \cite{zhong2025trajMatch0} minimize the discrepancy between full training trajectories. Distribution/feature matching methods \cite{liu2025distMatch0, li2025hdd} instead match statistical features or embedding distributions. Kernel-based methods\cite{chen2024kernal} formulate distillation as ridge regression under the Neural Tangent Kernel, enabling the derivation of analytically optimal distilled datasets. 

\par

In comparison, generative dataset distillation \cite{su2024diffusion, li2025diversity} methods leverage generative models to synthesize high-quality images with optimization objectives that benefit dataset distillation. The introduction of generative models decouples the distillation process and the generation process, reducing additional costs associated with increased IPC, enabling efficient generation of distilled datasets of flexible sizes \cite{wu2025dataset, ye2025information}. For instance, Li et al. \cite{li2024generative, li2025generative} introduce a GAN model and enhance the distillation performance by aligning the data distributions. Among generative models, diffusion models show great fidelity and robustness, bringing further advancement in dataset distillation. For example, D$^4$M \cite{su2024d4m} uses a latent diffusion model and achieves better generalizability across different architectures. CaO$_2$ \cite{wang2025cao2} proposes a two-stage framework that rectifies objective inconsistency and conditional mismatch in datasets obtained by diffusion models.

\par

Moreover, researchers continue to introduce new types of features to enhance the performance of dataset distillation from various perspectives, such as semantic information \cite{li2024semantic}. More diverse application scenarios have also been explored, such as recommendation system distillation \cite{zhang2025recommandDD}, multimodal distillation~\cite{li2025davdd}, and trustworthy dataset distillation \cite{ma2025PRtrustDD}. 

\section{Methodology}
\begin{figure}[p]
    \centering
    \begin{adjustbox}{rotate=90, center}
        \includegraphics[width=18cm]{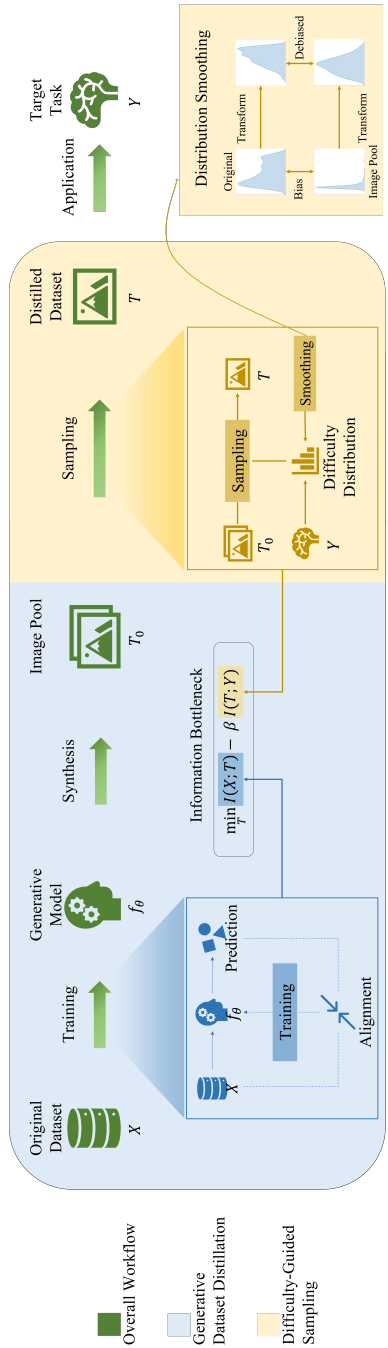}
    \end{adjustbox}
    \caption{Workflow of DGS. The overall processing is highlighted in green. The generation of the image pool is shown within the blue background. The sampling process is indicated within the yellow background. Distribution smoothing is applied to the difficulty distributions of both the original dataset and the image pool. The sampling distribution scales from the difficulty distribution of the original dataset.}
    \label{workflow}
\end{figure}

\begin{figure}[t]
    \centering
    \includegraphics[width=\linewidth]{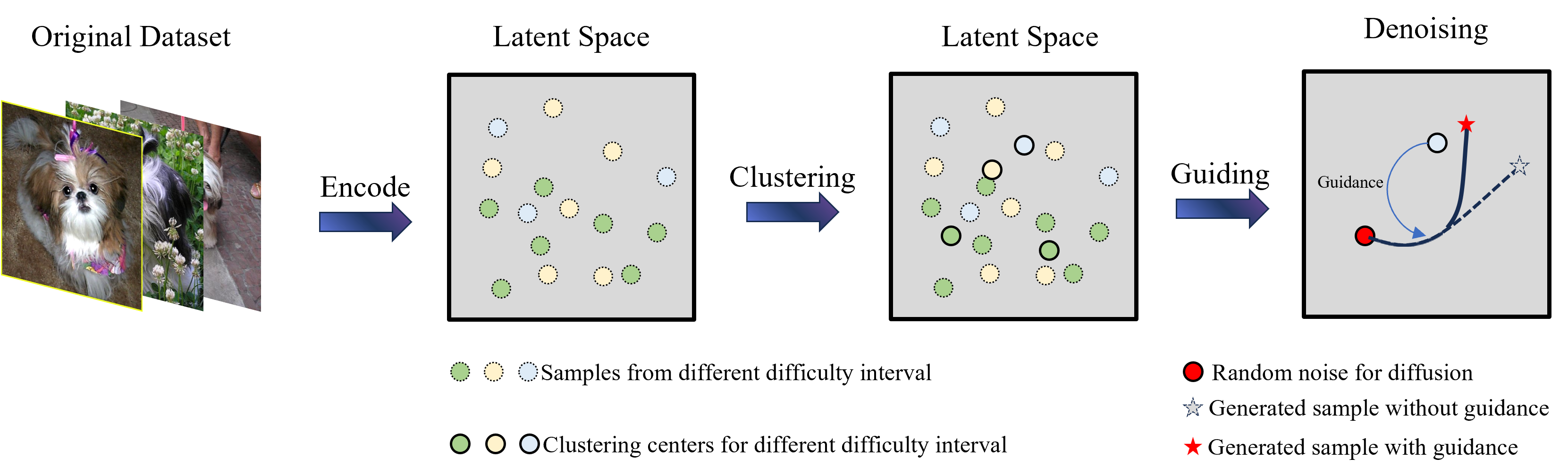}
    \caption{Workflow of DAG. The images in the original dataset are encoded into the latent space. Hierarchical clustering is conducted following the sampling distribution. Clustering centers are used to guide the denoising process.
    }
    \label{fig_dag}
\end{figure}

This section is organized as follows. First, we introduce a commonly adopted diffusion-based dataset distillation pipeline. Then we introduce the proposed DGS framework in detail, along with supporting theoretical analysis. Next, we describe the distribution smoothing strategy, designed to mitigate the bias in difficulty distribution between the original dataset and the image pool. Finally, we further explore the potential of difficulty as task-specific information from another perspective by proposing DAG. The workflows of DGS and DAG are illustrated in Fig.~\ref{workflow} and Fig.~\ref{fig_dag}, respectively. The detailed implementations are listed in Algorithm~\ref{alg_dgs} and Algorithm~\ref{alg_dag}.

\subsection{Preliminary}
\label{2-1}
To obtain high-quality image pools for sampling, we adopt the widely recognized Minimax \cite{gu2024minimax} as the baseline method, which proposes the Minimax criterion to improve the performance of Diffusion Transformer (DiT) \cite{Peebbles2023DiT} in dataset distillation. DiT operates in the latent space and combines the transformer architecture with the diffusion process, achieving a flexible and robust generation. The Minimax criterion is designed to enhance both the diversity and representativeness of the distilled datasets in a min–max manner during the fine-tuning of DiT.

\par

Given the original dataset $\mathcal{D} = {(\bm{x}_i, y_i)}$, to conduct processing in the latent space, the images are encoded into latent vectors using the VAE encoder. For simplicity, we assume a batch size of one during the training process. 
For the image $(\bm{x}_0, y_0)$ with the corresponding latent vector $\bm{z}_0$, the forward diffusion process gradually corrupts real data into noisy latent vector $\bm{z}_t$ by sequentially adding Gaussian noise $\epsilon \in \mathcal{N}(\bm{0}, \bm{I})$ to $\bm{z}_0$ over $t$ steps as follows:
\begin{equation}
    \bm{z}_t = \sqrt{\overline{\alpha}_t} \bm{z}_0 + \sqrt{1 - \overline{\alpha}_t} \epsilon, 
\end{equation}
where $\overline{\alpha}_t$ is a hyper-parameter called variance schedule. The diffusion model $f_{\theta}$, which is parameterized by $\theta$,  performs reverse denoising by predicting the added noise, thereby reconstructing the original image. The class information $\bm{c}$ is obtained by passing the class label $y_0$ through a class encoder. 
During training, the diffusion model learns to accurately reconstruct the added noise by minimizing the error between the predicted noise $f_{\theta}(\bm{z_t}, t, \bm{c})$ and ground-truth noise $\epsilon$ as follows:
\begin{equation}
    \mathcal{L}_{\text{diffusion}} = \arg \max_{\theta} {||f_{\theta}(\bm{z}_t, t, \bm{c}) - \epsilon||}_2^2.
\end{equation}
To fully leverage the advantages of diffusion models, a pre-trained model is typically used in dataset distillation. However, the diffusion-based training objective is insufficient for the goals of dataset distillation, necessitating additional strategies to fine-tune the model.

\par

The Minimax criterion aims to maximize both the representativeness and diversity of the distilled dataset, making better distribution alignment between the original dataset and the distilled dataset. To quantify these concepts, two auxiliary memory sets are constructed. The representativeness memory $\mathcal{M}_{r}$ and diversity memory $\mathcal{M}_{d}$ contain real images and generated images, respectively. Representativeness is calculated as the similarity between the original and the generated dataset. Improving representativeness is considered as putting close the generated and original image pair with the least similarity, i.e., the farthest distance, leading to the optimization objective as follows:
\begin{equation}
    \mathcal{L}_\text{repre} = \arg \max_{\theta} \min_{\bm{z}_{r} \in [\mathcal{M}_{r}]} \sigma(\bm{\hat{z}}_{\theta}(\bm{z}_t, \bm{c}), \bm{z}_{r}),
\end{equation}
where $\sigma(\text{·} \ , \ \text{·})$ means the cosine similarity between the two elements and $\bm{\hat{z}}_{\theta}(\bm{z}_t, \bm{c})$ is the predicted output of the diffusion model $f_{\theta}$ for noisy input $\bm{z}_t$ and class vector $\bm{c}$.
Similarly, diversity is estimated from the dissimilarity among the generated images. Improving diversity is considered as putting away the generated image pair with the most similarity,  i.e., the nearest distance, using the following optimization objective:
\begin{equation}
    \mathcal{L}_\text{div} = \arg \min_{\theta} \max_{\bm{\hat{z}}_{g} \in [\mathcal{M}_{d}]} \sigma(\bm{\hat{z}}_{\theta}(\bm{z}_t, \bm{c}), \bm{\hat{z}}_{g}).
\end{equation}
The diffusion loss $\mathcal{L}_{\text{diffusion}}$ helps the diffusion model to produce images that are similar to the original dataset, i.e., realistic images. And the minimax losses $\mathcal{L}_\text{repre}$ and $\mathcal{L}_\text{div}$ help improve the diversity and representativeness of the synthetic dataset. The integration of these two types of losses enables the diffusion model to generate datasets that better capture the nature of the original dataset, improving the performance of generated datasets in dataset distillation. The image pool $\mathcal{D}'$ is obtained using the fine-tuned model.

\par

Minimax essentially performs distribution alignment between the distilled and original datasets and achieves strong performance in image classification settings. However, similar to most generative dataset distillation methods, it overlooks information specific to the downstream task, like decision-relevant factors, inter-class relationships, or intra-class prioritization. The absence of such task-specific information leads to a target gap between the distillation process and the downstream task, limiting the optimal performance of the distilled datasets.

\subsection{Difficulty-guided Sampling}
\label{sec_3_2}
\begin{algorithm}[t]
    \caption{DGS}
    \label{alg_dgs}
    \begin{algorithmic}[1]
    \REQUIRE 
    $f_{\theta}$: an fine-tuned generative model parameterized by $\theta$;
    $f_{\text{cls}}$: a pre-trained image classification model;
    $\mathcal{D}=\{(\bm{x}, y)\}$: the original dataset;
    \ENSURE
    $\mathcal{D}^{\ast}$: the distilled dataset
    \STATE{Obtain the image pool $\mathcal{D}'$ using $f_{\theta}$}
        \FOR{ each dataset $\mathcal{D}_{0}$ in [$\mathcal{D}$, $\mathcal{D}'$]}
        \STATE{Estimate the difficulty distribution of the dataset $P_{\mathcal{D}_0}$ with $f_{\text{cls}}$ using Equation~\ref{equ_diff}}
        \STATE{Obtain threshold values $b_{\mathcal{D}_0}$ and $t_{\mathcal{D}_0}$ using Equation ~\ref{equ_th}}
        \STATE{Obtain the transformed distribution $P'_{\mathcal{D}_0}$ using Equation~\ref{equ_log}} and ~\ref{equ_clip}
        \ENDFOR
    \STATE{Obtain the sampling distribution $P_{\text{samp}}$ by scaling $P^*_{\mathcal{D}}$}
    \STATE{Sampling on $P'_{\mathcal{D}'}$ following $P_{\text{samp}}$}
    \STATE{Fetch the sampled images to form the distilled dataset $\mathcal{D}^{\ast}$}
    \end{algorithmic}
\end{algorithm}

We interpret dataset distillation from the perspective of the information bottleneck (IB) principle.
Let $X$ denote the original training dataset, $T$ denote the distilled dataset produced by a distillation algorithm, and $Y$ denote the label variable of the downstream task.
During downstream training, the learner only has access to $T$, not $X$, so $Y$ is conditionally independent of $X$ given $T$.
This induces the Markov chain $X \rightarrow T \rightarrow Y$, which makes it natural to view dataset distillation as an IB problem.

Under this view, an ideal distilled dataset $T$ should remove as much task-irrelevant information from $X$ as possible, while retaining information that is predictive of $Y$.
This trade-off can be formalized by the IB objective
\begin{equation}
  \mathcal{L}_{\mathrm{IB}} = \min_{T} I(X; T) - \beta I(T; Y),
  \label{eq:ib_objective}
\end{equation}
where $I(\cdot;\cdot)$ denotes mutual information and $\beta > 0$ is a Lagrange multiplier that balances compression and prediction.
Here, $I(X; T)$ measures how much information the distilled dataset keeps from the original data (including low-level appearance statistics and nuisance factors), while $I(T; Y)$ measures how informative the distilled dataset is for predicting the labels. The first term, therefore, encourages compressing the input to remove task-irrelevant information, and the second term encourages preserving task-relevant information. Balancing these two terms ensures that the distilled dataset remains compact while still supporting strong performance on the downstream task.

\par

Most existing generative distillation methods implicitly focus on surrogates of $I(X; T)$ by matching data distributions or feature statistics, but offer only weak control over $I(T; Y)$. For example, Minimax \cite{gu2024minimax} enhances the diversity and representativeness of the distilled dataset through distributional alignment, and RDED \cite{sun2024RDED} selects desired semantic parts as distilled datasets. Since the original dataset inherently contains diverse task-related information and serves as the ground truth, such approaches focus on $I(X; T)$, keeping as many features from the original dataset as possible in the fixed compression level. However, without explicit task-specific considerations on $I(T; Y)$, characteristics that contribute to solving the task remain constrained to the features in the original dataset. This shortness makes the distilled dataset suboptimal for a pre-determined downstream objective, especially when the task requires emphasizing particular features or relationships that are not sufficiently highlighted through keeping the original information.
 
\par

To address this limitation, we introduce task-specific information into dataset distillation to enhance $I(T; Y)$, ensuring that the distilled data contains sufficient information that contributes to the solution of $T$. This enables the distilled dataset to align more tightly with the target task and thereby achieve superior downstream performance. The works by Wang et al. \cite{wang2026difficulty} demonstrate that managing the difficulty of training samples benefits dataset quality in the field of dataset augmentation. Motivated by this insight, we use difficulty to represent the task-specific information in the image classification task and propose DGS to utilize this information towards improved performance of dataset distillation.

\par

Following the settings in the experiments of Wang et al., the inverse of the confidence $p$ assigned to the correct class $y_{true}$ is treated as the difficulty. A pre-trained classifier $f_{\mathrm{cls}}$ is introduced to estimate the confidence value. For an image $(\bm{x}, y_{true})$, its difficulty $d_{\bm{x}}$ is calculated as follows:
\begin{equation}
\label{equ_diff}
    d_{\bm{x}} = 1 - p_{f_{\text{cls}}}(y_{true} | \bm{x}).
\end{equation}
As a plug-in post-stage sampling module, DGS operates on an image pool generated by existing generative dataset distillation methods. Specifically, an image pool comprising $n \times \text{IPC}$ images is first constructed. We then compute the difficulty of the images using Equation~\ref{equ_diff}. The difficulty distribution of the image pool $P_{\mathcal{D}'}$ is used as additional annotations. Finally, the distilled dataset is sampled from the image pool according to a specific difficulty distribution $P_\text{samp}$.

\par

This sampling process integrates the task-specific information of difficulty into the distillation process. However, it also turns the critical factor in achieving good performance into the selection of an appropriate sampling distribution. This distribution should ideally capture the preferred difficulty composition of the training set for the classification task. Based on the premise that dataset distillation aims to retain the training performance of the original dataset, we hypothesize that datasets whose difficulty distribution approximates that of the original dataset may yield superior performance. To achieve this similarity, we partition the difficulty range into 10 intervals $I_k$ of width 0.1 and scale the original dataset’s difficulty distribution $P_\mathcal{D}$ to match the target IPC. In Section~\ref{sec_samp_distr}, we compare this scale-based sampling distribution with several pre-defined distributions, demonstrating the effectiveness of this sampling strategy.

\subsection{Distribution Smoothing}
However, as shown in Fig.~\ref{fig_kde}, distilled datasets produced by generative dataset distillation methods typically contain a disproportionately high proportion of easy samples. This preference introduces a clear bias in the difficulty distribution between the image pools and the original dataset. Such bias reduces the sampling coverage in certain difficulty intervals and causes the final sampled datasets to deviate from the desired difficulty distribution, thereby requiring additional corrective procedures.

\par

To address this issue, we apply distribution smoothing to better approximate the two distributions. The difficulty distributions of both the original dataset and the image pool are transformed toward a balanced state where they exhibit the best similarity. To achieve this, we introduce a modified logarithmic transformation equipped with a variable base and thresholding mechanism.

\par
First, the base of the transformation is determined dynamically by the minimum and maximum values of the dataset to keep the result between 0 and 1. For the difficulty distribution $P_{\mathcal{D}}(n)$, the logarithmic transformation $f$ is defined as follows:
\begin{equation}
\label{equ_log}
    f(P_{\mathcal{D}}, b, t)= \frac{\ln (P_{\mathcal{D}}(n) / \min(P_{\mathcal{D}}(n)))}{\ln (\max(P_{\mathcal{D}}(n)) / \min(P_{\mathcal{D}}(n)))}. 
\end{equation}

\par
However, simply using such a transformation leads to poor performance. This is because many samples in the dataset cluster around similar difficulty values, especially near the lower and upper extremes. Such concentration undermines the performance of the logarithmic transformation because it will amplify the influence of extreme values. To achieve better smoothing, we incorporate a thresholding mechanism that removes extreme values at both ends before transformation. The clipped distribution $P'_X(n)$ is obtained as follows:
\begin{equation}
\label{equ_clip}
    P'_{\mathcal{D}}(n)= H(n - b) \ P_{\mathcal{D}}(n) \ H(N - n - t) + \epsilon, 
\end{equation}
where $\epsilon$ is a small value to avoid mathematical error. $b$ denotes the bottom threshold and $t$ denotes the top threshold. $N$ is the number of samples in the dataset. And $H(n)$ is the Heaviside step function. The final transformation is applied to the clipped distribution by substituting $P'_{\mathcal{D}}(n)$ for $P_{\mathcal{D}}(n)$ in Equation~\ref{equ_log}.

\par

The thresholding mechanism operates by manually modifying extreme values, which inevitably introduces deviation from the original data. Therefore, the threshold values must be carefully selected to strike an appropriate balance between smoothing and deviation. To facilitate this selection, we introduce the Kullback-Leibler (KL) divergence to estimate distribution-level differences and determine the appropriate threshold values. For better smoothing, the ideal target is the uniform distribution $\mathcal{U}$, and the thresholding should maximize the similarity between $P'_{\mathcal{D}}(n)$ and $\mathcal{U}$. For less deviation, the ideal target is the original distribution $P_{\mathcal{D}}(n)$, and the thresholding should maximize the similarity between $P'_{\mathcal{D}}(n)$ and $P_{\mathcal{D}}(n)$. The selection of the two thresholds is formulated as follows:
\begin{align}
\begin{split}
\label{equ_th}
    b_{\mathcal{D}}, t_{\mathcal{D}} = \arg \min_{b, t} (\lambda \ D_\text{KL}(f(P_{\mathcal{D}}, b, t) || P_{\mathcal{D}}) + (1 - \lambda) \ D_\text{KL}(f(P_{\mathcal{D}}, b, t) || \mathcal{U})),
\end{split}
\end{align}
where $D_\text{KL}(P||Q)$ represents the KL divergence of P from Q. $\lambda \in [0, 1]$ is a weighting parameter deciding the degree of smoothing and deviation.

\par

Through these efforts, the sampling coverage of image pools can be improved, making the difficulty distribution of the distilled dataset better align with that of the original dataset. Working as an auxiliary component to DGS, distribution smoothing corrects the bias in difficulty distributions for image pools, thereby improving coverage of the sampling distribution and ensuring the faithful implementation of DGS.

\subsection{Difficulty-aware Guidance}
Although distribution smoothing helps to cover the sampling distribution better, it inevitably introduces new bias originating from the manipulation of the distribution. To solve the problem from scratch, reducing the dependence on smoothed distributions, we move beyond sampling and aim to directly generate datasets with the desired difficulty distribution instead. MGD3 \cite{chan-santiago2025mgd3} leverages class-wise clustering centers to guide the denoising process of the diffusion model, improving the distilled dataset's distributional coverage of the original dataset. Inspired by this work, we propose using difficulty centers as guidance to improve coverage on difficulty distribution. These centers serve to guide the denoising process toward samples that follow the difficulty pattern represented by the clustering centers. With this guidance mechanism, the distilled dataset inherently follows the intended difficulty distribution, allowing us to achieve the result of DGS from a different perspective.

\par

The division of difficulty intervals $I_k$ follows the same scheme as in DGS, with 10 intervals considered. Instead of clustering the entire dataset, K-means is performed separately on images within each interval $\bm{X}_k$. The number of clustering centers for each interval $I_{\text{samp}}(k)$ is determined according to the sampling distribution $I_{\text{samp}}$ used in DGS, which varies across different datasets and IPCs. For each clustering center $\bm{z}_{c}$ in the centers $\bm{Z}_{c}$, which is in the form of a latent vector, guidance is incorporated into the denoising process of a pre-trained diffusion model by adjusting the predicted latent vector $\bm{z}_t$ as follows: 
\begin{equation}
\label{equ_dag}
    \hat{\bm{z}_t} = \bm{z}_t + \lambda_{gui} (\bm{z}_{c} - \bm{z}_t) \sigma_t .
\end{equation}
where $\lambda_{gui}$ is a scalar parameter that determines the guidance strength and $\sigma_t$ is the variance schedule. Additionally, a stop timestep $t_{\text{stop}}$ is introduced to terminate the guidance at appropriate denoising steps to obtain better performance. DAG leverages difficulty-level clustering centers to capture common patterns shared by images in the same difficulty interval. These patterns are used to generate images of specific difficulty, filling the sampling distribution. Through this approach, the resulting generated datasets naturally align with the difficulty distribution of the original dataset, containing task-specific information.
\begin{algorithm}[t]
    \caption{DAG}
    \label{alg_dag}
    \begin{algorithmic}[1]
    \REQUIRE 
    $f_{\theta}$: an fine-tuned generative model parameterized by $\theta$;
    $f_{\text{cls}}$: a pre-trained image classification model;
    $\mathcal{D}=\{(\bm{x}, y)\}$: the original dataset;
    \ENSURE
    $\mathcal{D}^{\ast}$: the distilled dataset

    \STATE{Estimate the difficulty distribution of the original dataset $P_{\mathcal{D}}$ with $f_{\text{cls}}$ using Equation~\ref{equ_diff}}
    \STATE{Obtain the sampling distribution $P_{\text{samp}}$ by scaling  $P_{\mathcal{D}}$}
        \FOR{ each difficulty interval $I_k=\left[ k, k+0.1 \right], ~k~in~(0, 1, 0.1)$ }
        \STATE{Fetch images $\bm{X}_k$ of difficulty values in $I_k$}
        \STATE{Clustering $\bm{X}_k$ into $P_{\text{samp}}(k)$ centers.}
        \ENDFOR
        \FOR{ each center $\bm{z}_c$ in clustering centers $\bm{Z}_c$}
        \STATE{Generating an image using $f_{\theta}$ with guidance shown in Equation ~\ref{equ_dag}}
        \ENDFOR
    \STATE{Collect the generated images to form the distilled dataset $\mathcal{D}^{\ast}$}
    \end{algorithmic}
\end{algorithm}
\section{Experiments}
\label{sec_exp}

\subsection{Experimental Settings}
To validate the effectiveness of the proposed method, we conduct extensive experiments on different image pools across various experimental settings.  The original datasets include three 10-class subsets from the full-size ImageNet dataset: ImageWoof \cite{fastai2019imageNette}, ImageNette \cite{fastai2019imageNette}, and ImageIDC \cite{kim2022IDC}. ImageWoof consists of 10 specific dog breeds and is known as a challenging dataset for image classification. ImageNette contains 10 classes that are easy to classify, and ImageIDC consists of 10 classes that are randomly selected from ImageNet. The downstream models include ConvNet-6 \cite{gidaris2018convNet}, ResNet-18 \cite{he2016resNetAP}, and ResNet-10 with average pooling (ResNetAP-10) \cite{he2016resNetAP}, with a learning rate of 0.01. Distilled datasets are used to train the classification model from scratch, and top-1 accuracy scores on the validation set are recorded for comparison. Each experiment is conducted 3 times, with the mean value and standard deviation presented.

\par

First, we conduct DGS on image pools obtained by various baseline methods. Then, we validate the robustness of DGS by repeating the experiments with different downstream models and across various datasets. We also show visual results on difficulty distributions and sampled images to offer an intuitive understanding. Then, we show the performance on different selections of hyperparameters to determine the appropriate value. Finally, we validate the effectiveness of DAG by conducting experiments on ImageNette and comparing the results with the baseline method MGD3. 

\par

\begin{sidewaystable}[p]
    \centering
    \caption{Comparison of downstream validation accuracy with various baseline methods using different downstream models. The experiments are conducted on ImageWoof. DGS (Ori) represents performing DGS on the original dataset. DiT/Minimax + DGS represents performing DGS on image pools generated by DiT/Minimax. The highest accuracy values in each comparison are highlighted in bold.}
    \label{exp_main_gen}
    \begin{tabularx}{\linewidth}{c|c|cXc|cc|cX|cX|c}
        \hline
        IPC (Ratio) & Test Model & Random & K-Center \cite{sener2018kcenter} & DGS (Ori) & DiT \cite{Peebbles2023DiT} & DiT + DGS & Minimax \cite{gu2024minimax} & Minimax + DGS & Full Dataset 
        \\ 
        \hline
        & ConvNet-6 & $24.3_{\pm 1.1}$ & $19.4_{\pm 0.9}$ & \bm{$24.6_{\pm 0.7}$} & \bm{$31.4_{\pm 0.4}$} & $31.0_{\pm 1.1}$ & $34.1_{\pm 0.4}$ & \bm{$35.2_{\pm 0.4}$} & $86.4_{\pm 0.2}$ 
        \\ 
        10 (0.8\%) & ResNetAP-10 & $28.4_{\pm 0.7}$ & $22.1_{\pm 0.1}$ & \bm{$28.8_{\pm 0.9}$} & $34.4_{\pm 0.6}$ & \bm{$35.1_{\pm 1.0}$} & $35.7_{\pm 0.3}$ & \bm{$37.8_{\pm 0.3}$} & $87.5_{\pm 0.5}$ 
        \\
        & ResNet-18 & $27.7_{\pm 0.9}$ & $21.1_{\pm 0.4}$ & \bm{$28.0_{\pm 0.7}$} & $33.9_{\pm 0.1}$ & \bm{$34.2_{\pm 0.2}$} & $35.3_{\pm 0.4}$ & \bm{$36.0_{\pm 0.4}$} & $89.3_{\pm 1.2}$ 
        \\ 
        \hline
         
        & ConvNet-6 & $29.1_{\pm 0.7}$ & $21.5_{\pm 0.8}$ & \bm{$31.2_{\pm 0.4}$} & $
        36.9_{\pm 1.1}$ & \bm{$37.2_{\pm 0.3}$} & $36.9_{\pm 1.2}$ & \bm{$39.5_{\pm 0.6}$} & $86.4_{\pm 0.2}$ 
        \\ 
        20 (1.6\%) & ResNetAP-10 & $32.7_{\pm 0.4}$ & $25.1_{\pm 0.7}$ & \bm{$37.4_{\pm 0.7}$} & $41.1_{\pm 0.8}$ & \bm{$41.5_{\pm 0.7}$} & $43.3_{\pm 0.3}$ & \bm{$45.4_{\pm 0.4}$} & $87.5_{\pm 0.5}$ 
        \\ 
        & ResNet-18 & $29.7_{\pm 0.5}$ & $23.6_{\pm 0.3}$ & \bm{$33.3_{\pm 0.5}$} & $40.5_{\pm 0.5}$ & \bm{$41.0_{\pm 0.4}$} & $40.9_{\pm 0.6}$ & \bm{$42.4_{\pm 0.4}$} & $89.3_{\pm 1.2}$ 
        \\ 
        \hline
         
        & ConvNet-6 & $41.3_{\pm 0.6}$ & $36.5_{\pm 1.0}$ & \bm{$41.6_{\pm 0.6}$} & $46.5_{\pm 0.8}$ & \bm{$47.1_{\pm 1.4}$} & $51.4_{\pm 0.4}$ & \bm{$51.5_{\pm 0.2}$} & $86.4_{\pm 0.2}$ 
        \\ 
        50 (3.8\%) & ResNetAP-10 & $47.2_{\pm 1.3}$ & $40.6_{\pm 0.4}$ & \bm{$48.7_{\pm 0.2}$} & $49.3_{\pm 0.2}$ & \bm{$52.5_{\pm 0.7}$} & $54.4_{\pm 0.6}$ & \bm{$57.1_{\pm 0.9 }$} & $87.5_{\pm 0.5}$ 
        \\ 
        & ResNet-18 & $47.9_{\pm 1.8}$ & $39.6_{\pm 1.0}$ & \bm{$47.7_{\pm 0.2}$} & $50.1_{\pm 0.5}$ & \bm{$53.5_{\pm 1.5}$} & $53.9_{\pm 0.6}$ & \bm{$54.3_{\pm 0.8}$} & $89.3_{\pm 1.2}$ 
        \\
        \hline
    \end{tabularx}
\end{sidewaystable}

For Minimax, we adopt the official configuration provided by the authors. During the distillation process, a pre-trained DiT with Difffit \cite{xie2023Difffit} is fine-tuned for 8 epochs over the full dataset using a batch size of 8. Input images are randomly shuffled, augmented, and resized to $256 \times 256$ before being encoded into the latent space through the VAE encoder. The optimization is performed with AdamW at a learning rate of 1e-3. The generation stage consists of 50 denoising steps. A ResNet-50 model trained on the complete ImageNet dataset serves as the difficulty estimator. The thresholding weight $\lambda$ in distribution smoothing is set to 0.5.

\subsection{Benchmark Results}
\begin{table}
    \centering
    \footnotesize
    \renewcommand{\arraystretch}{1.5}
    \caption{Comparison of downstream validation accuracy with various dataset distillation methods. The results are reported on ImageWoof using ResNet-18. The highest accuracy values are highlighted in bold.}
    \begin{tabularx}{\linewidth}{Y|YYY}
        \hline
        Method &  IPC=10 & IPC=20 & IPC=50
        \\
        \hline
        Random & $27.7_{\pm 0.9}$ & $29.7_{\pm 0.5}$ & $47.9_{\pm 1.8}$
        \\
        K-Center \cite{sener2018kcenter} & $21.1_{\pm 0.4}$ & $23.6_{\pm 0.3}$ & $39.6_{\pm 1.0}$
        \\
        Herding \cite{Welling2009Herding} & $30.2_{\pm 1.2}$ & $32.2_{\pm 0.6}$ & $48.3_{\pm 1.2}$
        \\
        \hline
        DM \cite{zhao2023DM} & $33.4_{\pm 0.7}$ & $29.8_{\pm 1.7}$ & $46.2_{\pm 0.6}$
        \\
        IDC-1 \cite{kim2022IDC} & \bm{$36.9_{\pm 0.4}$} & $38.6_{\pm 0.2}$ & $48.3_{\pm 0.8}$
        \\
        D4M \cite{su2024d4m} & $33.2_{\pm 2.1}$ & $40.1_{\pm 1.6}$ & $51.7_{\pm 3.2}$
        \\
        DiT \cite{Peebbles2023DiT} & $34.7_{\pm 0.4}$ & $40.5_{\pm 0.5}$ & $50.1_{\pm 0.5}$
        \\
        Minimax \cite{gu2024minimax} & $35.3_{\pm 0.4}$ & $40.9_{\pm 0.6}$ & $53.9_{\pm 0.6}$
        \\
        \hline
        DGS & $36.0_{\pm 0.4}$ & \bm{$42.4_{\pm 0.4}$} & \bm{$54.3_{\pm 0.8}$}
        \\
        \hline
    \end{tabularx}
    \label{exp_sota}
\end{table}

We conduct experiments on the ImageWoof with various IPC settings and different downstream models to show the fundamental performance of DGS. 
First, we directly sample on the original dataset and compare the results with dataset selection methods like Random and K-Center \cite{sener2018kcenter}. Then, we conduct DGS on image pools generated by DiT and Minimax, and compare the performance between DGS and the baseline methods. As shown in Table~\ref{exp_main_gen}, DGS consistently achieves improved accuracy than baseline methods in most experiments, especially in high IPC settings. The results validate its ability to enhance the task-specific performance of current dataset distillation methods as a plug-in module. Because the combination of DGS and Minimax achieves current SOTA performance, the following experiments use Minimax as the default baseline method unless otherwise specified, marked directly as DGS. We also list the performances of more dataset distillation methods in Table~\ref{exp_sota}, including selection methods like Herding \cite{Welling2009Herding}, and dataset distillation methods like DM \cite{zhao2023DM}, IDC-1 \cite{kim2022IDC}, and D$^4$M \cite{su2024d4m}.

\par

Then, we verify the generalization ability of DGS by expanding the experiments to more datasets. As shown in Table~\ref{exp_dataset}, the performance of DGS on ImageNette and ImageIDC generally corresponds with the trend on ImageWoof, with DGS showing improved performance than Minimax and achieving the best performance in all the experiments. 

\subsection{Visualization}
\begin{figure*}[t]
        \centering
        \includegraphics[width=\linewidth]{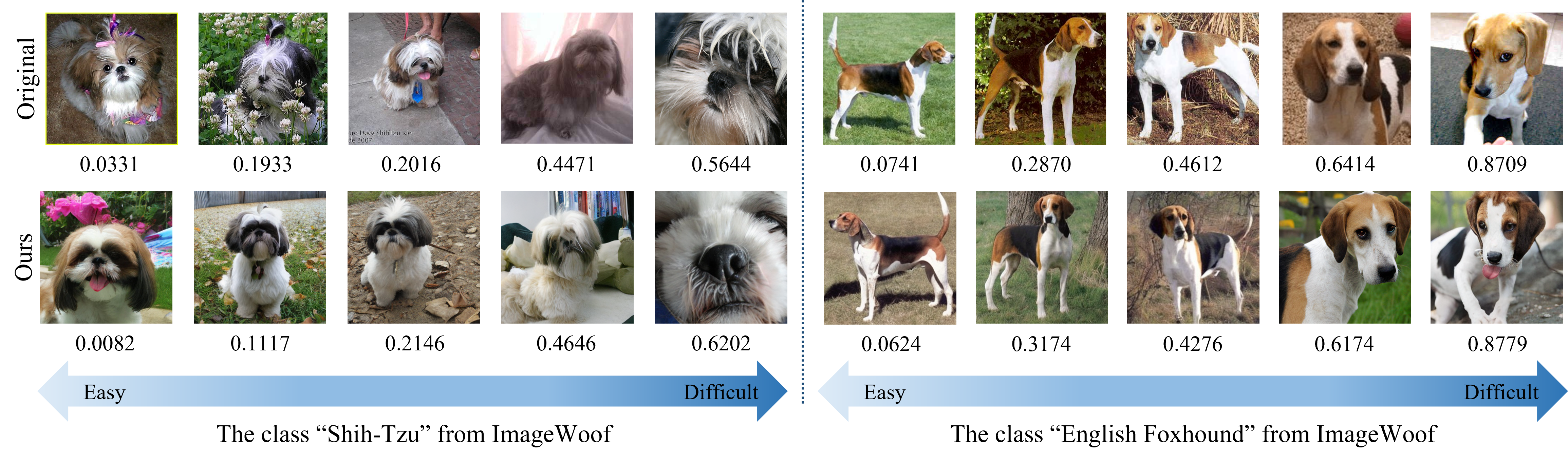}
        \caption{Images from the original dataset and the DGS-sampled dataset with difficulty scores. The sampled dataset follows the difficulty distribution of the original dataset. Images with similar difficulty scores exhibit shared characteristics.}
        \label{fig_visual}
\end{figure*}

\begin{figure}[p]
    \centering
    \begin{adjustbox}{rotate=90, center}
        \includegraphics[width=18cm]{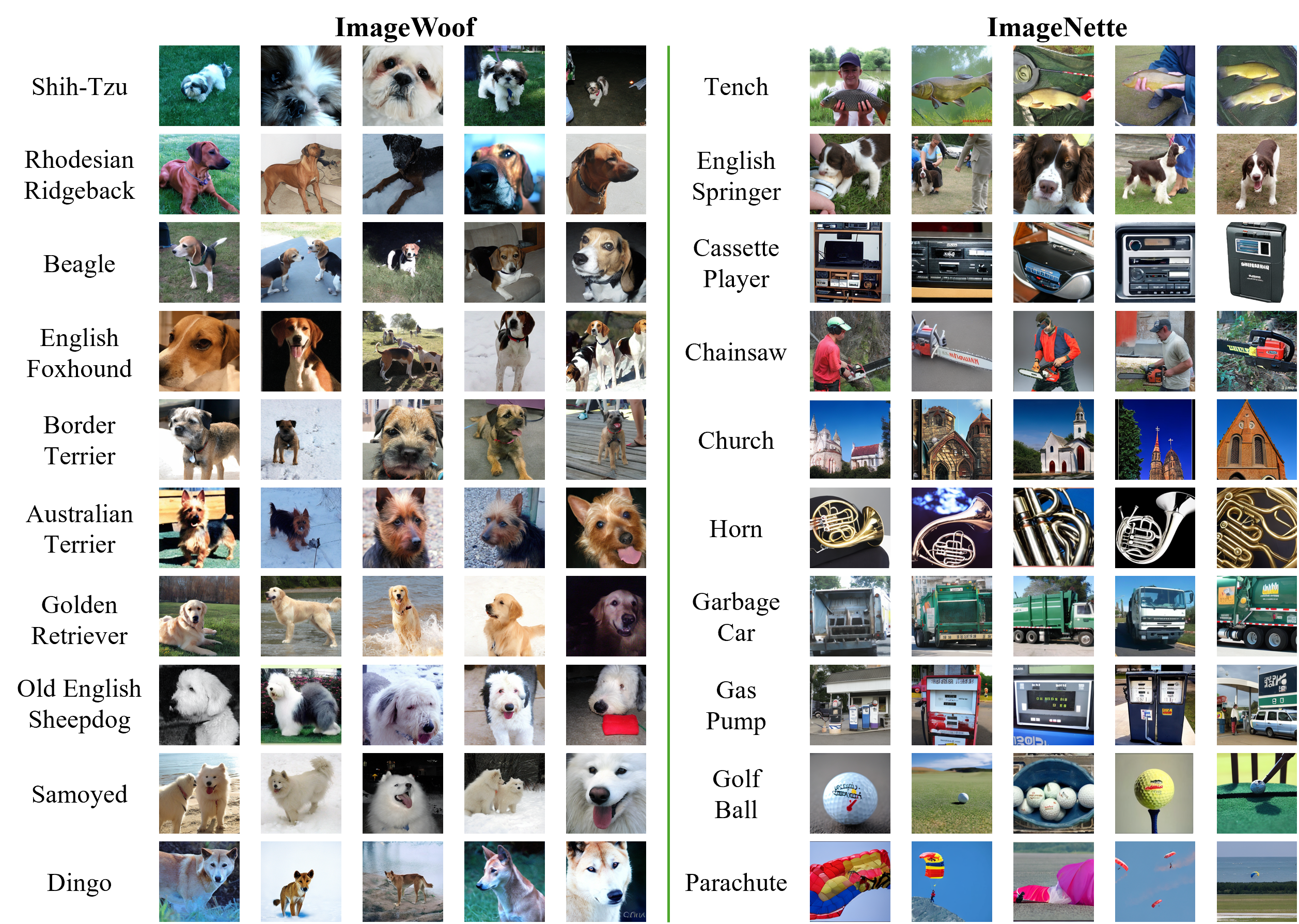}
    \end{adjustbox}
    \caption{Images from various classes in the distilled datasets obtained by DGS on ImageWoof and ImageNette.}
    \label{fig_disImgAll}
\end{figure}

In Fig.~\ref{fig_visual}, we present visual samples from the two aforementioned classes together with their difficulty values. The results show that sampled datasets include images spanning a broad range of difficulty levels, similar to the original dataset. Moreover, images with similar difficulty scores tend to exhibit shared visual characteristics, suggesting that DGS implicitly promotes characteristic diversity within the distilled dataset. We also present additional samples from several classes across ImageWoof and ImageNette in Fig.~\ref{fig_disImgAll}, offering a qualitative overview of the datasets sampled by DGS.
\begin{figure*}[t]
        \centering
        \includegraphics[width=\linewidth]{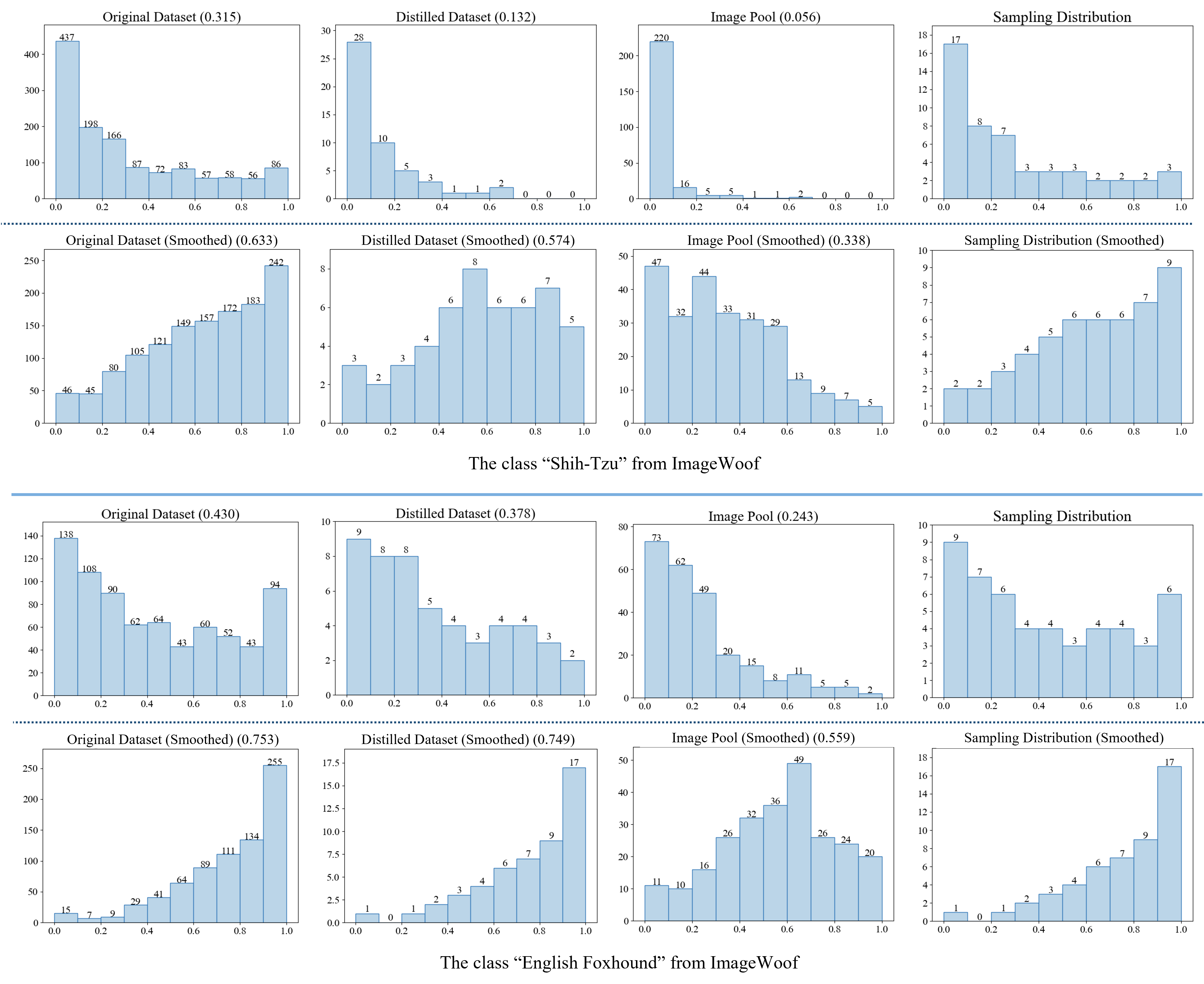}
        \caption{Illustration of difficulty distributions during DGS of two classes from ImageWoof. The horizontal axis shows difficulty intervals, while the vertical axis shows the number of samples in each interval. The average difficulty is indicated in the title. The first row contains original distributions, and the second row presents those after distribution smoothing.}
        \label{fig_distr}
\end{figure*}

\begin{table}
    \centering
    \footnotesize
    \renewcommand{\arraystretch}{1.5}
    \caption{Comparison of downstream validation accuracy with other baseline methods on various ImageNet subsets. Results are reported using the downstream architecture of ResNetAP-10. The highest accuracy values are highlighted in bold.}
    \begin{tabularx}{\linewidth}{c|P{1.5cm}|YYYY}
        \hline

        & IPC & Random & DiT \cite{Peebbles2023DiT} & Minimax \cite{gu2024minimax} & DGS \\

        \hline
        \multirow{3}*{\rotatebox{90}{ImageNette}} & 10 & $54.2_{\pm 1.6}$  & $59.1_{\pm 0.7}$ & $59.8_{\pm 0.3}$ & \bm{$61.5_{\pm 0.9}$}
        \\
        & 20 & $63.5_{\pm 0.5}$ & $64.8_{\pm 1.2}$ & $66.3_{\pm 0.4}$ & \bm{$66.9_{\pm 0.5}$}
        \\
        & 50 & $76.1_{\pm 1.1}$ & $73.3_{\pm 0.9}$ & $75.2_{\pm 0.2}$ & \bm{$76.8_{\pm 0.7}$} 
        \\

        \hline
        \multirow{3}*{\rotatebox{90}{ImageIDC}} & 10  & $48.1_{\pm 0.8}$ & $54.1_{\pm 0.4}$ & $60.3_{\pm 1.0}$ & \bm{$61.6_{\pm 0.7}$}
        \\
        & 20 & $52.5_{\pm 0.9}$ & $58.9_{\pm 0.2}$ & $63.9_{\pm 0.4}$ & \bm{$64.3_{\pm 0.5}$}
        \\
        & 50 & $68.1_{\pm 0.7}$ & $64.3_{\pm 0.6}$ & $74.1_{\pm 0.2}$ & \bm{$74.2_{\pm 0.7}$}
        \\
        \hline
    \end{tabularx}
    \label{exp_dataset}
\end{table}
\par

To offer an intuitive view of how DGS operates, we visualize the difficulty distributions at different stages during the sampling process, using the ImageWoof classes ``Shih-Tzu" (n02086240) and ``English Foxhound" (n02089973) as examples. The two examples show different original dataset difficulty distributions, but both can be handled properly by distribution smoothing.

\par

As shown in the first row of Fig.~\ref{fig_distr}, the image pools exhibit a strong preference toward easy samples, resulting in bias to the original dataset. This bias also causes deviations from the sampling distribution shown in the fourth column, which is designed to follow the original distribution. Consequently, sampled datasets have several difficulty intervals that remain sparsely populated or even empty. Although random selection can be applied to fill these gaps, doing so weakens the effect of DGS and pushes the sampling process closer to random selection. In contrast, the second row of Fig.~\ref{fig_distr} demonstrates that the logarithmic transformation effectively smoothens the difficulty distributions, eliminating the discrepancy between the original dataset and the image pool. This enhances the coverage of the sampling distribution, ensuring the faithful implementation of DGS.  

\subsection{Sampling Distribution}
\label{sec_samp_distr}
\begin{figure}[t]
    \centering
    \includegraphics[width=\linewidth]{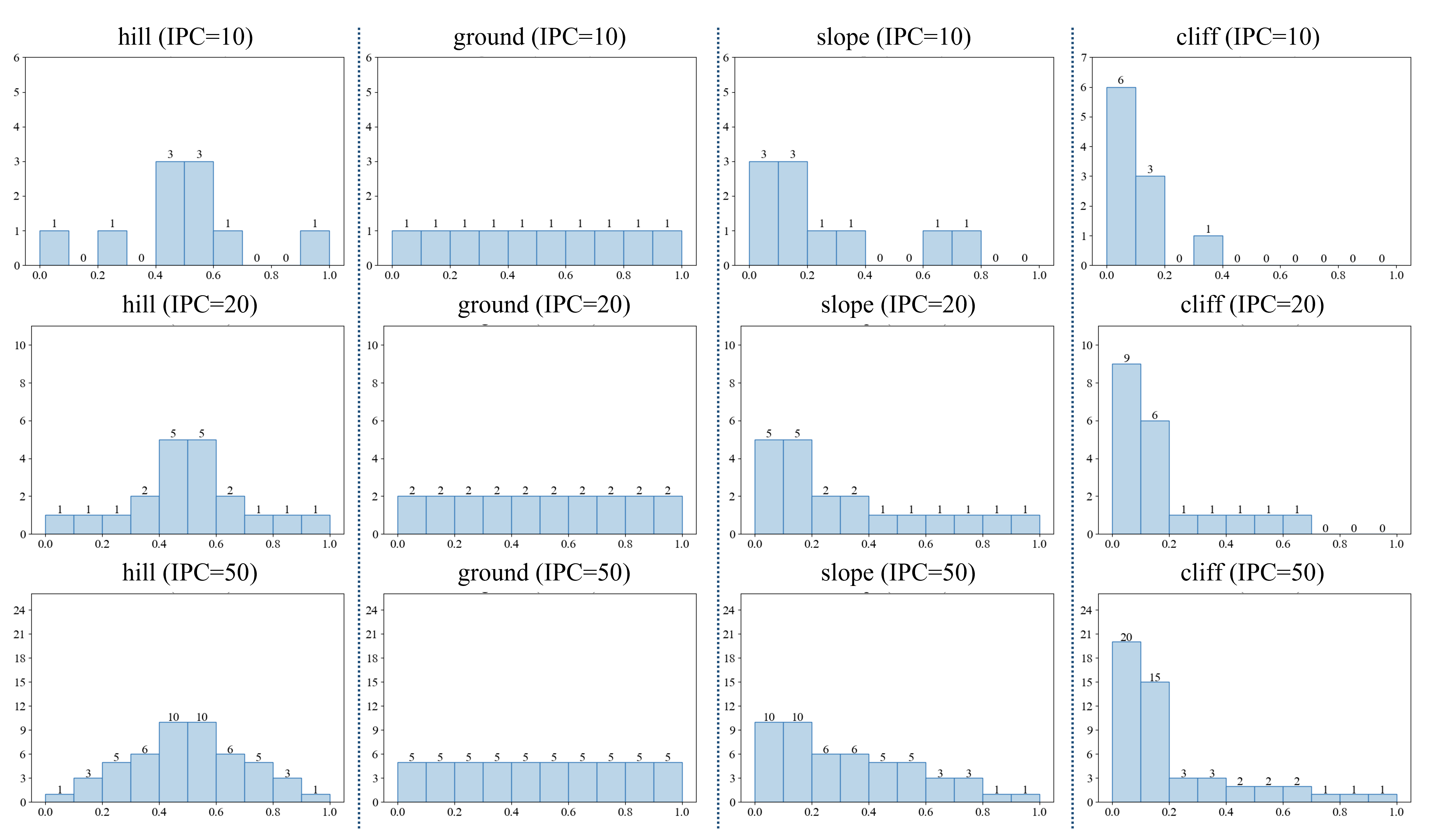}
    \caption{Visualization of pre-defined sampling distributions across various IPC configurations. 
    From left to right, the distributions have increasingly larger ratios of easy examples.
    The horizontal axis corresponds to difficulty intervals, while the vertical axis denotes the target number of samples within each interval.
    }
    \label{fig_samp_distr}
\end{figure}

We hypothesize in Sec.\ref{sec_3_2} that datasets whose difficulty distribution approximates that of the original dataset may yield superior performance and propose the scale-based sampling distribution. To validate this hypothesis, we compare the performance of scale-based sampling distribution with several pre-defined sampling distributions illustrated in Fig.~\ref{fig_samp_distr}. These predefined distributions differ in the ratios of easy samples and are named based on their visual shapes. 

\par

The results in Table~\ref{exp_samp_distr} indicate that the scale-based sampling distribution achieves the best overall performance. Additional analysis of the results across the sampling distributions shows a trend that when the size of the sampled dataset is small, allocating more easy samples may benefit the training, whereas in larger sampled datasets, increasing the ratio of difficult samples leads to better performance. This observation indicates that different IPC settings demonstrate unique preferences for difficulty distribution, offering new insights for adjusting the sampling distribution according to each specific IPC setting.

\subsection{Size of Image Pool}
Since final distilled datasets are sampled from the image pool, the pool size may substantially affect the downstream performance. To quantify this impact and determine a practical configuration, we evaluate image pools of multiple sizes, which are scaled to the IPC as $n \times \text{IPC}$. The experimental results summarized in Table~\ref{exp_ip_size} show that the validation accuracy fluctuates with the size. 
Based primarily on the performance observed in high IPCs, a pool size of $5 \times \text{IPC}$ provides the most consistent gains and is thus used as the default value. This observation reflects an inherent trade-off: increasing the pool size enhances sample diversity, but excessively large pools tend to accumulate redundant samples, particularly when difficulty values cluster within narrow ranges. Additionally, the pool size interacts with the selection of threshold parameters in the distribution smoothing, resulting in different sampling distributions.

\begin{table}
    \centering
    \footnotesize
    \renewcommand{\arraystretch}{1.5}
    \caption{Comparison of downstream validation accuracy for different sampling distributions. The label ``scale" refers to the sampling distribution scaled from the difficulty distribution of the original dataset. The experiments are conducted on ImageWoof using ResNetAP-10. The highest accuracy values are highlighted in bold.}
    \centering
    \begin{tabularx}{\linewidth}{Y|YYY}
        \hline
        Distribution &  IPC = 10 & IPC = 20 & IPC = 50
        \\
        \hline
        Hill & $37.7_{\pm 0.2}$ & $42.7_{\pm 0.7}$ & $56.1_{\pm 0.6}$ 
        \\
        Ground & $35.8_{\pm 0.2}$ & $41.7_{\pm 0.3}$ & $56.9_{\pm 0.5}$
        \\
        Slope & $37.4_{\pm 0.6}$ & $44.3_{\pm 0.3}$ & $56.6_{\pm 0.8}$
        \\
        Cliff & $36.7_{\pm 0.7}$ & $42.7_{\pm 0.8}$ & $55.0_{\pm 0.6}$
        \\
        Scale & \bm{$37.8_{\pm 0.3}$} & \bm{$45.5_{\pm 0.4}$} & \bm{$57.1_{\pm 0.9}$}
        \\
        \hline
    \end{tabularx}
    \label{exp_samp_distr}
\end{table}
\begin{table}
    \centering
    \footnotesize
    \renewcommand{\arraystretch}{1.5}
    \caption{Comparison of downstream validation accuracy across image pools of various sizes. The experiments are conducted on ImageWoof using ResNetAP-10. The highest accuracy values are highlighted in bold.}
    \centering
    \begin{tabularx}{\linewidth}{Y|YYY}
        \hline
        Size &  IPC=10 & IPC=20 & IPC=50
        \\
        \hline
        2 $\times$ IPC & $37.9_{\pm 0.4}$ & $44.3_{\pm 0.3}$ & $55.8_{\pm 0.6}$
        \\
        3 $\times$ IPC & $35.9_{\pm 0.8}$ & $41.5_{\pm 0.5}$ & $56.7_{\pm 0.4}$
        \\
        4 $\times$ IPC & \bm{$38.4_{\pm 0.6}$} & $43.7_{\pm 0.5}$ & $55.4_{\pm 0.9}$
        \\
        5 $\times$ IPC & $37.4_{\pm 0.3}$ & \bm{$45.5_{\pm 0.4}$} & \bm{$57.1_{\pm 0.9}$}
        \\
        6 $\times$ IPC & $37.8_{\pm 0.3}$ & $42.7_{\pm 1.1}$ & $54.9_{\pm 1.3}$
        \\
        \hline
    \end{tabularx}
    \label{exp_ip_size}
\end{table}

\subsection{Validation Sets}

During our experiments, we observed that prior works typically use a single fixed validation set, which may be insufficient to reflect practical deployment scenarios. To more comprehensively evaluate the robustness of DGS across different downstream settings, we construct multiple validation sets from the original dataset, which show different difficulty preferences. We perform DGS in datasets generated by Minimax and compare their performance across these validation sets. 

\par
\begin{table}
    \footnotesize
    \renewcommand{\arraystretch}{1.5}
    \caption{Comparison of downstream validation accuracy on different validation sets. The label ``scale'' represents the validation set sampled from the original dataset using the sampling distribution of DGS. The results are obtained with ResNetAP-10 on ImageNette. The best results for each validation set are marked in bold.}
    \centering
    \begin{tabularx}{\linewidth}{c|P{2cm}|YYY}
        \hline
        Set & Method & IPC=10 & IPC=20 & IPC=50
        \\
        \hline
        Ori & Minimax & $59.8_{\pm 0.3}$ & $66.3_{\pm 0.4}$ & $75.2_{\pm 0.2}$
        \\
        & DGS & \bm{$61.5_{\pm 0.9}$} & \bm{$66.9_{\pm 0.5}$} & \bm{$76.8_{\pm 0.7}$}
        \\
        \hline
        Random & Minimax & $57.1_{\pm 0.3}$ & $63.5_{\pm 0.4}$ & $72.2_{\pm 0.3}$ 
        \\
        & DGS & \bm{$59.7_{\pm 0.3}$} & \bm{$66.0_{\pm 0.5}$} & \bm{$75.9_{\pm 0.3}$}
        \\
        \hline
        Easy & Minimax & $60.3_{\pm 1.1}$ & $67.6_{\pm 0.5}$ & \bm{$81.2_{\pm 0.7}$} 
        \\
        & DGS & \bm{$62.3_{\pm 0.4}$} & \bm{$71.5_{\pm 0.5}$} & $79.5_{\pm 0.3}$
        \\
        \hline
        Scale & Minimax & $56.9_{\pm 0.8}$ & $62.6_{\pm 0.6}$ & $73.8_{\pm 0.7}$
        \\
        & DGS & \bm{$59.9_{\pm 0.7}$} & \bm{$64.5_{\pm 0.2}$} & \bm{$74.1_{\pm 0.5}$}
        \\
        \hline
    \end{tabularx}
    \label{exp_vali}
\end{table}

As shown in Table~\ref{exp_vali}, although the performances of both the methods fluctuate in different validation sets, distilled datasets sampled by DGS consistently show improved performance with better stability across different validation sets.

\subsection{Difficulty-aware Guidance}
We propose DAG as an alternative utilization strategy of difficulty, where distilled datasets of desired difficulty distribution are directly generated rather than being sampled. We show the performance of DAG across various $t_{stop}$ in Table~\ref{exp_DAG} and compare it with the baseline method MGD3 \cite{chan-santiago2025mgd3}. The stop guidance of MGD3 is set to 25, which is the best value mentioned in the original paper. The results show that with an appropriate stop timestep, DAG consistently improves the performance of MGD3. Although DAG is not a plug-in sampling module like DGS, its performance highlights the broader potential of incorporating task-specific information using various implementation strategies.
\begin{table}
    \centering
    \footnotesize
    \renewcommand{\arraystretch}{1.5}
    \caption{Comparison of downstream validation accuracy between DAG and MGD3. The results are obtained with ResNetAP-10 on ImageNette. The highest accuracy values are highlighted in bold.}
    \centering
    \begin{tabularx}{\linewidth}{Y|YYY}
        \hline
        Method-$t_{stop}$ & IPC=10 & IPC=20 & IPC=50
        \\
        \hline
        MGD3-25 & $64.9_{\pm 0.4}$ & $71.1_{\pm 0.5}$ & $79.3_{\pm 0.2}$
        \\
        DAG-25 & \bm{$65.6_{\pm 0.3}$} & \bm{$72.3_{\pm 0.2}$} & \bm{$80.3_{\pm 0.4}$}
        \\
        DAG-30 & $65.1_{\pm 0.4}$ & $67.7_{\pm 0.7}$ & $78.6_{\pm 0.2}$
        \\
        DAG-40 & $63.6_{\pm 1.4}$ & $67.5_{\pm 0.9}$ & $76.3_{\pm 0.4}$
        \\
        DAG-50 & $61.4_{\pm 0.8}$ & $67.1_{\pm 0.4}$ & $70.6_{\pm 0.3}$
        \\
        \hline
    \end{tabularx}
    \label{exp_DAG}
\end{table}

\section{Conclusion}
In this paper, we investigate the role of task-specific information in bridging the target gap between dataset distillation and the downstream task. For image classification, we propose DGS to align the difficulty distributions between the distill and the original dataset by means of post-stage sampling on image pools obtained by SOTA methods. In addition, we propose DAG to discuss the potential of difficulty in the generation process. Both methods achieve improved performance over baseline approaches in most experiments, confirming the utility of difficulty for the classification task. 
Despite these improvements, several directions remain open for further advancement. Experiments on datasets from diverse data domains like SVHN can further explore the generalizability of the proposed methods. The combination of current difficulty with other characteristics, like classification uncertainty, may serve as a more credible definition. And other forms of informative signals, including semantic similarity, can be incorporated for cooperation. 
Beyond image classification, this work also provides a general perspective for incorporating various forms of task-specific information to enhance the performance of dataset distillation on different downstream tasks, such as confidence maps for object detection. The information can provide quantitative values to perform sampling or optimization that benefits the downstream training. 

\section*{Acknowledgments}
This study was supported in part by JSPS KAKENHI Grant Numbers JP23K21676, JP24K02942, JP24K23849, and JP25K21218.

\bibliographystyle{elsarticle-num}
\bibliography{PR}
\end{document}